\documentclass{article}

\usepackage[final, nonatbib]{neurips_2023}

\usepackage[utf8]{inputenc} 
\usepackage[T1]{fontenc}    
\usepackage{hyperref}       
\usepackage{url}            
\usepackage{booktabs}       
\usepackage{amsfonts}       
\usepackage{nicefrac}       
\usepackage{microtype}      
\usepackage{xcolor}         
\usepackage{graphicx} 
\usepackage{epsfig} 
\usepackage{mathptmx} 
\usepackage{times} 
\usepackage{amsmath} 
\usepackage{amssymb}  
\usepackage{csquotes}
\usepackage{flushend}
\usepackage{algorithm}
\usepackage{algpseudocode}
\usepackage{csquotes}
\usepackage{indentfirst}
\usepackage{caption}
\usepackage{subcaption}

\usepackage{makecell}
\usepackage{color}
\usepackage{overpic}
\usepackage{soul}
\usepackage{threeparttable}
\usepackage{hyperref}

\usepackage{listings}
\usepackage[most]{tcolorbox}
\tcbuselibrary{breakable}

\newtcolorbox{codebox}[1][]{
  colback=gray!20,
  colframe=gray!60,
  rounded corners,
  listing only,
  listing options={basicstyle=\ttfamily, breaklines=true},
  title=#1
}

\title{
Interactive Planning Using Large Language Models for Partially Observable Robotics Tasks
}

\author{%
  Lingfeng Sun\\
  University of California, Berkeley\\
  \texttt{lingfengsun@berkeley.edu} \\
  \And
  Devesh K. Jha \\
  Mitsubishi Electric Research Laboratories \\
  \texttt{jha@merl.com} \\
  \AND
  Chiori Hori \\
  Mitsubishi Electric Research Laboratories \\
  \texttt{chori@merl.com} \\
  \And
  Siddarth Jain \\
  Mitsubishi Electric Research Laboratories \\
  \texttt{sjain@merl.com} \\
  \And
  Radu Corcodel \\
  Mitsubishi Electric Research Laboratories \\
  \texttt{corcodel@merl.com} \\
  \And
  Xinghao Zhu\\
  University of California, Berkeley\\
  \texttt{zhuxh@berkeley.edu} \\
  \And
  Masayoshi Tomizuka\\
  University of California, Berkeley\\
  \texttt{tomizuka@berkeley.edu} \\
  \And
  Diego Romeres \\
  Mitsubishi Electric Research Laboratories \\
  \texttt{romeres@merl.com} \\
}

\begin{document}

\maketitle

\begin{abstract}
Designing robotic agents to perform open vocabulary tasks has been the long-standing goal in robotics and AI. Recently, Large Language Models (LLMs) have achieved impressive results in creating robotic agents for performing open vocabulary tasks. However, planning for these tasks in the presence of uncertainties is challenging as it requires \enquote{chain-of-thought} reasoning, aggregating information from the environment, updating state estimates, and generating actions based on the updated state estimates. In this paper, we present an interactive planning technique for partially observable tasks using LLMs. In the proposed method, an LLM is used to collect missing information from the environment using a robot and infer the state of the underlying problem from collected observations while guiding the robot to perform the required actions. We also use a fine-tuned Llama 2 model via self-instruct and compare its performance against a pre-trained LLM like GPT-4. Results are demonstrated on several tasks in simulation as well as real-world environments. A video describing our work along with some results could be found \href{https://youtu.be/9vJmqVti8to}{here}. 
\end{abstract}

\section{Introduction}
Designing robots that have the physical intelligence to perform open vocabulary tasks is extremely challenging. This requires that robots be able to interpret tasks from an open set of instructions and execute them robustly while performing the required reasoning. One can argue that this could be the most challenging problem facing artificial intelligence (AI). However, designing such agents can truly revolutionize the way robots would be integrated into our future society. Recently, large language models (LLMs)~\cite{openai2023gpt4, anil2023palm, touvron2023llama2} have been shown to be very impressive at solving tasks of different complexities~\cite{vemprala2023chatgpt, saycan,yu2023language,li2022pretrained,liang2023code}. 
Large language models can help understand the tasks and decompose them into a sequence of actions, reward functions, or goals for policy given appropriate prompts and training data. 
Motivated by these developments, we present a problem of interactive planning in uncertain environments where a robot may not have complete information to perform the task. In these tasks, the robot needs to interact with its environment and collect additional information to complete the task.

\begin{figure}[h]
	\centering
        \includegraphics[width=1\textwidth]{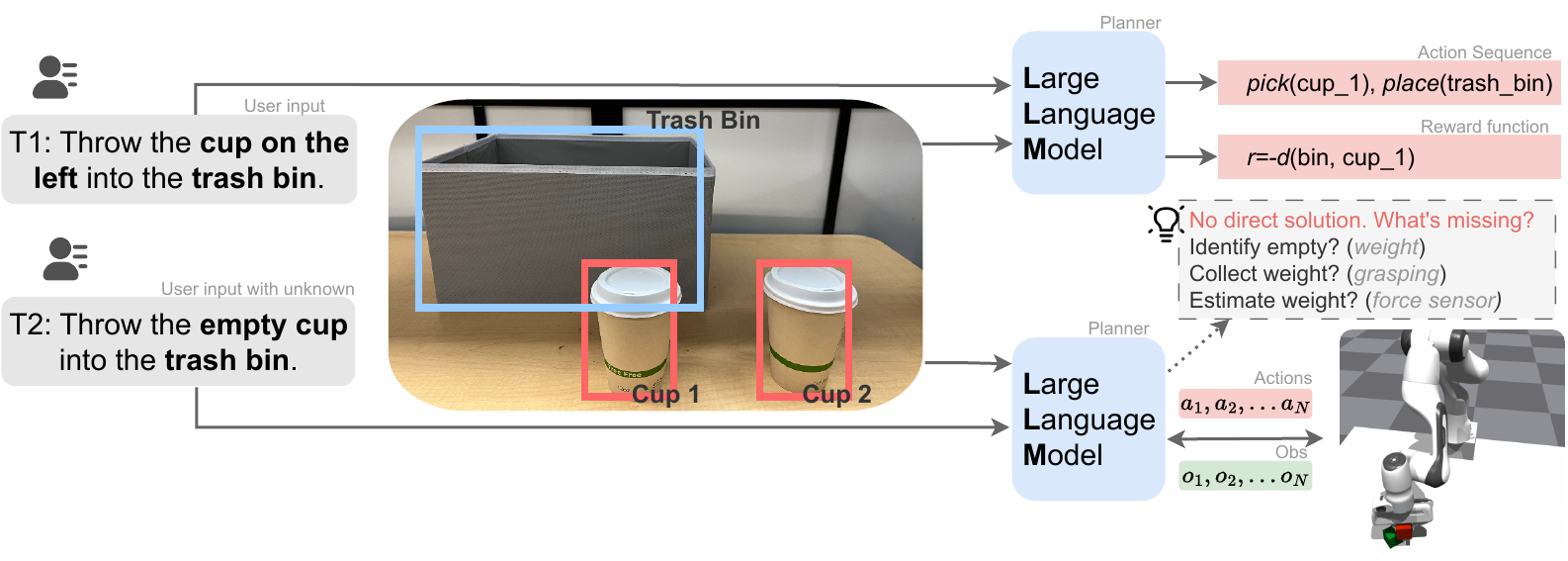}
	\vspace{-0.1in}
	\caption{An example task where the uncertainty is present in the content of the cups. For task T1, the robot is asked to throw the cup on the left into the bin. An LLM agent can generate feasible action sequences for the robot to perform the task. When asked to throw the empty cup, the agent cannot reason which cup is empty based on current information. It needs to interact with the cups and use feedback from observations (e.g., force sensor reading) to identify the empty cup. }
	\vspace{-0.15in}
	\label{fig:example_tasks}
\end{figure}

Partial observability and uncertainty are the norm, rather than the exception, in the real world. For example, consider task T2 shown in Figure~\ref{fig:example_tasks}, where a robot needs to understand how it can gather information to identify the empty cup and then throw it in the bin. Unlike the tasks with complete information, it would be challenging to design a sequence of skills or a suitable reward function that can solve this task. This problem can be formulated as a Partially Observable Markov Decision Process (POMDP)\cite{pomdp}. However, solving POMDPs could be computationally intractable. It requires reasoning in the belief state of the problem and does not scale well with the dimensionality of the problem. Prior work on using LLMs for robotic tasks has demonstrated good reasoning capability of LLMs as well as mapping of the reasoning to robot actions~\cite{saycan,huang2022inner,driess2023palme}. Inspired by these advancements, we believe that we can leverage the reasoning and \enquote{chain-of-thoughts}(CoT) capability of LLMs to solve partially observable tasks while interacting with the environment. What makes this challenging for current LLMs is the requirement to understand real robot observations from different modalities and use them for task planning.

Most of the prior works using LLMs in robotics focused on step-wise scene and task understanding making full use of the current available modalities to infer the optimal action and/or reward~\cite{saycan,huang2022inner,yu2023language, driess2023palme}. In this work, we focus on performing interactive planning under cases of partial observability. This requires planning to aggregate information from the environment, reasoning about the correct state of the system, and updating the state estimates based on the sensor observations collected by the robot. Furthermore, we also try to understand how well a fine-tuned smaller model like Llama2-7B \cite{touvron2023llama2} performs in comparison with a pre-trained LLM like GPT-4. The smaller models are generally desirable for practical reasons but it could be challenging to distill the reasoning capability of models like GPT-4 for complex robotic tasks discussed in this paper. To understand this, we propose an instruction data generation pipeline following the self-instruction~\cite{selfinstruct} scheme to understand the limitations of smaller models and potential ways to overcome them.

In summary, our work makes the following contributions:
\begin{itemize}
\item We introduce the Large Language Model for Partially Observable Task Planning(LLM-POP) framework to interactively plan with uncertainties. We demonstrate the framework in simulation and real-world environments.
\item We compare the performance of pre-trained LLM as well as fine-tuned smaller models in the proposed framework for partially observable tasks. 
\end{itemize}


\section{Related Work}

\noindent\textbf{LLMs for embodied AI tasks:}
The recent strides in the field of LLMs have renovated the robotics community in various task domains, especially on task planning \cite{brohan2023rt1, vemprala2023chatgpt,wang2023voyager,li2022pretrained,sharma2022skill}, where the robot is asked to reason about language instructions to generate robot actions, reward functions for online controllers \cite{yu2023language}, and code\cite{ liang2023code}. Previous works also combine LLMs with Task and Motion Planning (TAMP) to make use of the traditional motion planning algorithms \cite{ding2023task,lin2023text2motion,chen2023autotamp}. To solve robotics tasks in larger and more complex settings, researchers have utilized LLMs to process multi-modal information, \cite{jansen2020visuallygrounded,huang2023voxposer,saycan,driess2023palme}, enable multi-robot \cite{mandi2023roco} and human-robot collaboration \cite{cui2023onlinecorrection}. 
A series of works use feedback in the planning process~\cite{saycan, huang2022inner, yao2022react, sharma2022correcting, cui2023onlinecorrection} to improve the LLM-generated plans, and use the natural language provided by the user to correct robot behaviors. Other works model the uncertainties in tasks~\cite{ren2023robots, chen2023asking} and gather reasoning by involving human knowledge in the decision loop. In our problem setting, we focus on the partial observation environment setting where initial information is insufficient for task solving, and the robot actively seeks task-relevant information via sensory feedback. 

\noindent\textbf{LLMs for Data Synthesis and instruction tuning:}
Recent open-sourced models like LLaMa \cite{touvron2023llama}\cite{touvron2023llama2}, Alpaca \cite{alpaca}, and Gorrila \cite{patil2023gorilla}, explore how pre-trained models can be improved for general and specific tasks \cite{shen2023hugginggpt, qin2023toolllm, tang2023toolalpaca, yang2023autogpt}. 
Researchers have proposed strategies to prompt LLMs to perform high-quality synthetic data by in-context learning\cite{kojima2023large}, task decomposition\cite{shridhar2023distilling, yao2022react}, as well as methods to improve fine-tuning performance \cite{hu2021lora, gao2023llamaadapter}, and pre-train model for specific tasks like coding \cite{rozière2023code}.
Our research incorporates these methods to get synthetic data for instruction-tuning but focuses on tasks that require reasoning, but with limited labeled data. 
\section{Interactive LLM Planning with Uncertainties}

The objective of the proposed framework is to perform long-horizon robotic tasks in the presence of various kinds of uncertainties using LLMs. These tasks require a closed-loop, interactive planning where the robot should be able to collect useful observations from the environment and then make optimal decisions. An example of such a task is illustrated in Figure~\ref{fig:example_tasks} where the robot's task is to throw the empty cup into the bin. However there exists uncertainty in the contents of the cups, and therefore this information needs to be obtained by sensorimotor operations and provided as the feedback to the LLM. 
For clarity of presentation, this section delves into the formulation of the underlying problem using the notion of POMDPs. We then elaborate on the pivotal role of LLMs in the interactive planning framework.


\subsection{Problem Formulation}
\subsubsection{\textbf{Partial Observation setting}}
A POMDP is an extension of a traditional MDP that tackles decision-making scenarios where the agent lacks complete state information. A POMDP is defined by a tuple \((S, A, P, R, \Omega, O)\), with \(\Omega\) as the observation set and O as the observation function.  At each time step, the environment is in state $s\in S$. The agent takes action $a\in A$ and causes the environment to transit to $s'$ accordingly to the transition function $P(s'|s, a)$. At the same time step, the agent gets an observation $o\in \Omega $ which depends on the current state of the environment $O(o|s')$.
Unlike the policy function in MDP $\pi(a|s)$, which maps the underlying states to the actions, POMDP's policy $\pi(a|b)$ is a mapping from the belief states $b$ to the actions. The belief state $b$ is a probabilistic estimation of the full state $s$. The updated belief state $b'$ after observing $o$ is described by: 
$b'(s') = C\cdot O(o|s')\sum_{s\in S} P(s'|s, a)$ where $C$ is a normalizing constant.

We also want the proposed framework to be generalizable to a variety of tasks. For different tasks $\tau$, the information required to make decisions can differ. This adds additional complexity since now the LLM has to reason about a generalizable state space $S$. In the open-vocabulary robotics task scenarios, the robot observations are determined by on-board sensors. Not all information about the environment is relevant to the task; some of them can be directly extracted from observations, while some are unknown and require exploration. Thus, we end up getting task-dependent belief state $b^{\tau}$, and the task-related states $s^{\tau}$ for task $\tau$. Both finding the necessary state abstraction for different tasks and finding the optimal policy $\pi$ under the task-specific MDP is important in this task-dependent POMDP setting. 


\subsubsection{\textbf{Action space of robots}}
For long-horizon tasks, using a pre-trained set of parameterized skills as action space is a common choice. In this paper, we use a set of parameterized skills like \{\textbf{pick, place, reach, reset}\}. All these skills can be performed using robot observations and thus we do not consider partial observability during robot skills execution. It is noted that we do not consider continuous sensory feedback during skill execution-- however, that could be incorporated by training skills using RL. 

\subsubsection{\textbf{Uncertainties in Tasks}}
The uncertainty in decision-making in the tasks we test mainly arises from two aspects:
 
\noindent\textbf{Environmental Uncertainty:} These uncertainties arise in the POMDP settings due to the agent's lack of complete environmental knowledge. For example, physical properties of the objects that cannot be directly observed. The uncertainties in the belief $b^{\tau}$ can be reduced with certain observations. This is a major challenge we target to solve in this paper. 

\noindent\textbf{Skill Execution Uncertainty:} Even with a well-defined plan, the actual execution of actions on robots might not always lead to the expected outcome. This can be mainly attributed to the difference between the transition functions $P$, $P_{\textrm{real}}$ of the designed and real system as well as unexpected disturbances during execution.

With the challenges explained above, we propose a framework where LLMs are used as policy as well as for state abstraction for the underlying POMDP. 

\subsection{Language-based Planners}
\label{llm_planner}
Based on the problems described in the previous section, we propose to use a LLM to play a multifaceted role in the interactive planning process:

\noindent \textbf{LLM for State Abstraction:} Given the environment description and sensor observations, LLM needs to analyze the available information and abstract sufficient statistics (or the appropriate state) to solve the task. Furthermore, it needs to reason about what is uncertain based on the current observations. It needs to update its belief based on the observations when prompted with historical information.

\noindent \textbf{LLM as Policy:} Given the observation and action space, LLM needs to plan actions that gather environmental information to mitigate the uncertainty and update the agent's belief state. The LLM-based policy is also expected to generate the optimal plan to maximize the reward based on the task description with minimal steps. Also, since we use open-loop parameterized skills for the robot, the LLM is also used to provide feedback to the robot in cases of failure in the execution of these skills. This feedback needs to be provided in a way that is still executable by the robot.

We use LLM to reason about these problems during task execution. It is noted that actions in the POMDP setting is conditioned on new observations and updated beliefs. There are a few additional challenges when using LLM as closed-loop policy for tasks with uncertainties that we consider in the paper. To update the belief state of the task, the LLM must understand the robot observations from different modalities (pose detections, force sensors, etc.). These data formats might be new to the LLM model and thus, must be properly included in the prompt template to the LLM. Furthermore, the skills available to the robot are parameterized by continuous position and orientation coordinates which might be challenging to reason about while performing robotic tasks. Similarly, the output of the language model needs to be executable by the robot; the response should be written in a template that the downstream controller can understand. In the next section, we will discuss how we use the LLMs to solve the interactive planning task.


\section{LLM-POP: Interactive Planning}
\begin{figure}[t]
    \centering
    \includegraphics[width=10cm]{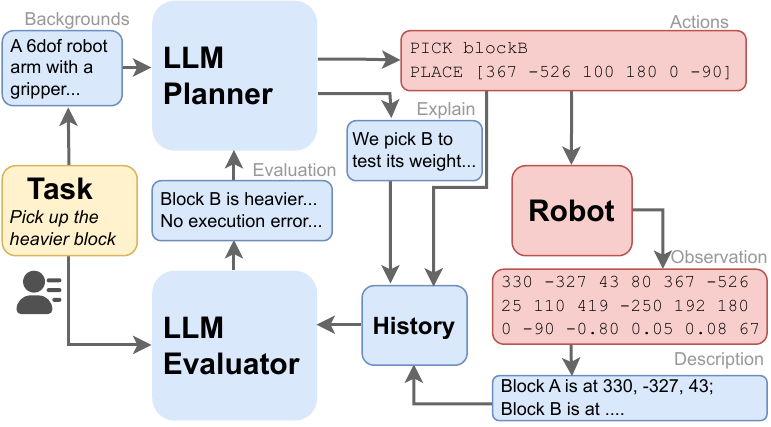}
    \vspace{-0.1in}
    \caption{An example showing how the framework works during solving the task ``Pick up the heavier block''. The LLM planner outputs an executable action sequence to the robot. The robot executes the action and the observation description and action pair is added into history buffer. The LLM evaluator analyzes the historical information and outputs the updated information into the planner to generate a new plan.}
    \label{fig:framework}
    \vspace{-0.2in}
\end{figure}
The proposed framework (LLM-POP) for interactive planning is illustrated in Figure~\ref{fig:framework}. As introduced in the problem formulation, the language-based policy in our framework has multiple tasks to do in the planning loop. 
At each step, the input to the language model contains the \textbf{task description} from a user, \textbf{the current observation} from the robot, and the \textbf{historical action and observation sequence} from previous steps. The model output includes an executable \textbf{sequence of actions} and the \textbf{corresponding text explanation}. The robot will execute the actions provided by the policy output and return the observations for a next-round query of the LLM. The language model must finish the reasoning task and output the policies in the designed format. The task description is the only user-provided input during the planning process. 
In the following sections, we show how we use a pre-trained LLM (GPT-4) as well as a fine-tuned smaller model to serve as the planner and evaluator. 
\subsection{Prompt structure for GPT-4}\label{sec:prompt_gpt4}
Using powerful LLMs like GPT-4 as interactive planners relies on its strong chain-of-thought reasoning and in-context learning capability. Therefore, the prompt (input of a single round LLM query) to the LLM requires careful design to ensure it can generalize to robotics tasks and avoid hallucination (generating actions in wrong formats or not executable for the robot) in responses.

As shown in Figure \ref{fig:planprompt}, the prompt template for the \textit{planner} consists of the following parts: 
\begin{itemize}
    \item \textit{Environment description, action options, output rules}: Background information that help understand the task settings. This information is preset by the user and is constant throughout planning for different tasks. 
    \item \textit{Task description}: texts describing tasks from users. We assume the first two parts should provide enough information for the LLM to understand what's the missing information and what are actions that can collect the information.
    \item \textit{Example outputs}: in-context examples for planning.
    \item \textit{Current observation and historical information}: text-format descriptions of current observation and historic information. If the observation is poses and force, use vectors with explanations.
\end{itemize}

The explanation in output, together with the action sequence, will be included in historical information. This helps the LLM to understand the past actions it has performed and avoid reasoning about it again. Note that the LLM planner needs to specify the parameters in the actions based on its own understanding of the environment, task, and the action space description. For manipulation tasks, this includes location and orientation for the target pose.

As shown in Figure \ref{fig:framework}, along with the LLM \textit{planner}, we also designed an LLM \textit{evaluator} using a similar prompt structure. The evaluator also takes in the background information, task description, and history observations after executing past actions. It evaluates the task-execution status and appends it to next-round prompting. As described in Section \ref{llm_planner}, the evaluator here will explicitly ask the LLM to finish the ``state abstraction'' (analyze what's the missing information), ``belief update'' in policy (analyze information from historical observations), and ``correct execution errors''(identify failures from the history).
Although it is possible to put all the requirements into the LLM planner, asking it to do all the analysis and make planning decisions in the response, we find decomposing this into two steps improves the reasoning results. Example prompts of planner and evaluator are shown in Section \ref{appendix:prompt_planner}, \ref{appendix:prompt_evaluator}.

\subsection{Fine-tuning a smaller model as planner}
\label{sec:fine-tune}
Fine-tuning a language model, rather than directly querying a GPT-4, not only enables offline deployment but also holds distinct advantages in the context of interactive planning. One prominent reason is the incorporation of multi-modality in the data. Our system doesn't solely rely on text descriptions but also utilizes the robot's observations. While these observations can theoretically be converted into text form, they constitute a novel data type that GPT-4 has not been trained on, thereby resulting in limited zero-shot generalizability. 
For example, in experiments using GPT-4, if poses in robot observations and action parameters are in different frames of reference, the LLM will have trouble transforming them.
A second reason is the requirement of large contexts in the input. A direct query to GPT would necessitate the inclusion of environment settings and generation constraints at each instance, which is inefficient and cost-intensive.
\begin{figure}[t]
    \centering
    \begin{subfigure}[t]{0.45\textwidth}
         \centering
         \includegraphics[width=\textwidth]{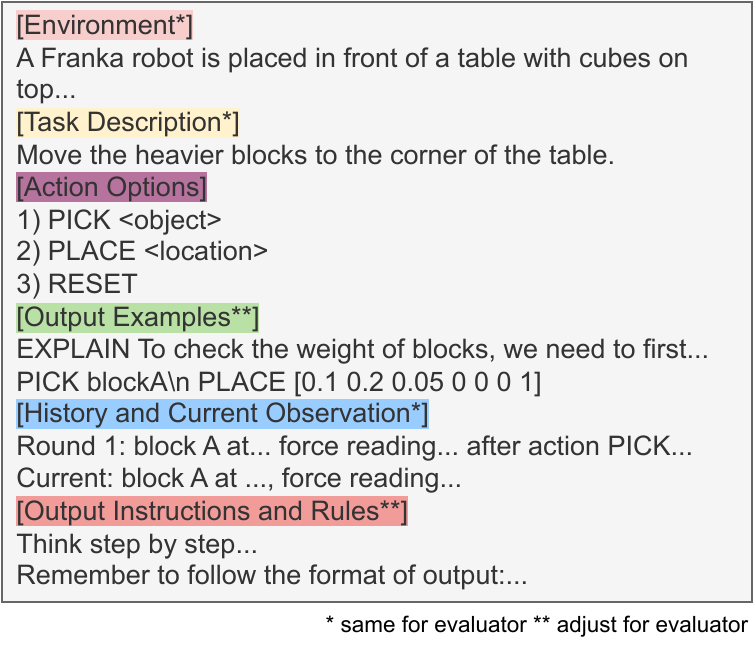}
         \caption{Prompt template for GPT planner and evaluator. The task description is taken from user input, the others are pre-defined according to the environment and robot. Output rules will be adjusted for the evaluator.}
         \label{fig:planprompt}
     \end{subfigure}
     \hfill
     \begin{subfigure}[t]{0.50\textwidth}
         \centering
         \includegraphics[width=\textwidth]{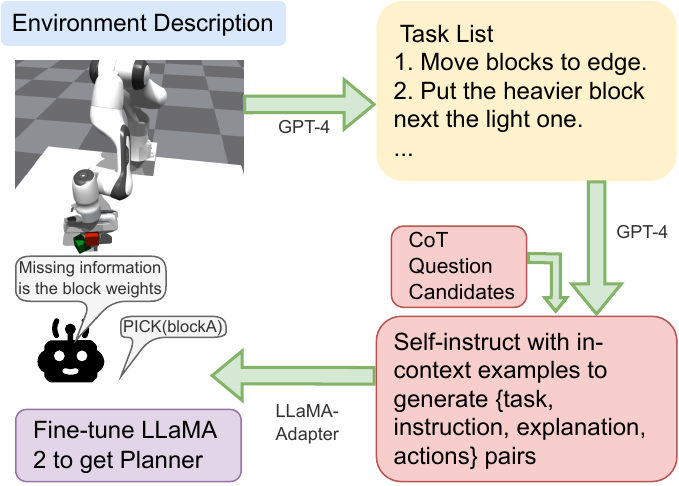}
         \caption{The training procedure of the fine-tuned LLMs as an interactive planner as described in Sec \ref{sec:fine-tune}. During inference, questions come from the pre-defined CoT question set, inputs come from robot observation.}
         \label{fig:self_instruct}
     \end{subfigure}
    \vspace{-0.1in}
    \caption{}
\end{figure}
The difficulty of fine-tuning a smaller pre-trained LLM model mainly comes from two sides: 
1) \textbf{Lack of data} for complex tasks. Most robotics data in the wild\cite{walke2023bridgedata,roboagent,fang2023rh20t} has no partial observable tasks involved, and force-torque sensor data is usually not included since they are noisy and vary across robots.
2) Smaller models are \textbf{worse at reasoning tasks}, CoT is tied with larger models~\cite{cot}.

In order to get the required data to fine-tune a model as a planner in interactive planning under partial observation, we follow the procedure shown in Figure \ref{fig:self_instruct}, using self-instruct\cite{selfinstruct} to generate an instruction dataset and fine-tune a LLaMA2-7B\cite{touvron2023llama} model. The full pipeline includes:

\noindent\textbf{Task Generation}: The description of the environment, robot, potential uncertainties, action options, and example tasks are provided to  GPT-4 to generate a number of tasks that are feasible to solve. We encourage GPT-4 to make the task set diverse in difficulty. An example prompt template and examples of generated tasks are shown in Section \ref{appendix:prompt_task}, \ref{appendix:example_task}.

\noindent\textbf{Instruction Generation}: The generated tasks are used to generate pairs of instructions and responses, following the self-instruct paradigm. The instruction includes task descriptions and questions, the input encompasses the robot's observations. The output generated by the model includes the same verbal explanations and actions as GPT-4 planners. We add format instructions to guarantee the ``response'' format. An example prompt template and examples of generated instructions are shown in Section \ref{appendix:prompt_instruct}, \ref{appendix:example_instruct}.

\noindent\textbf{CoT question designs}: Finishing the state abstraction, belief update, and action planning in one query is hard for smaller models. Therefore, we create CoT questions\cite{ho2023large} to ask \textit{if missing information exists}, \textit{how to collect information}, and \textit{how to solve the task with fill information}. The planner will choose questions to ask based on binary options in response.

\noindent\textbf{Integrating collected robot observations}: For the pre-trained actions, we collect success trajectories of the robot finish the actions and use them as in-context reference examples in the \textit{Instruction Generation} process.

\noindent\textbf{Fine-tuning}: For the fine-tuning process, we adopted the LLaMA-adapter \cite{gao2023llamaadapter}. This approach allows us to enhance the model's performance by leveraging a specifically curated dataset and fine-tuning it to our unique task generation and interactive planning scenario. 

\section{Experiments}
\begin{figure}[h!]
    \centering
    \includegraphics[width=1\textwidth]{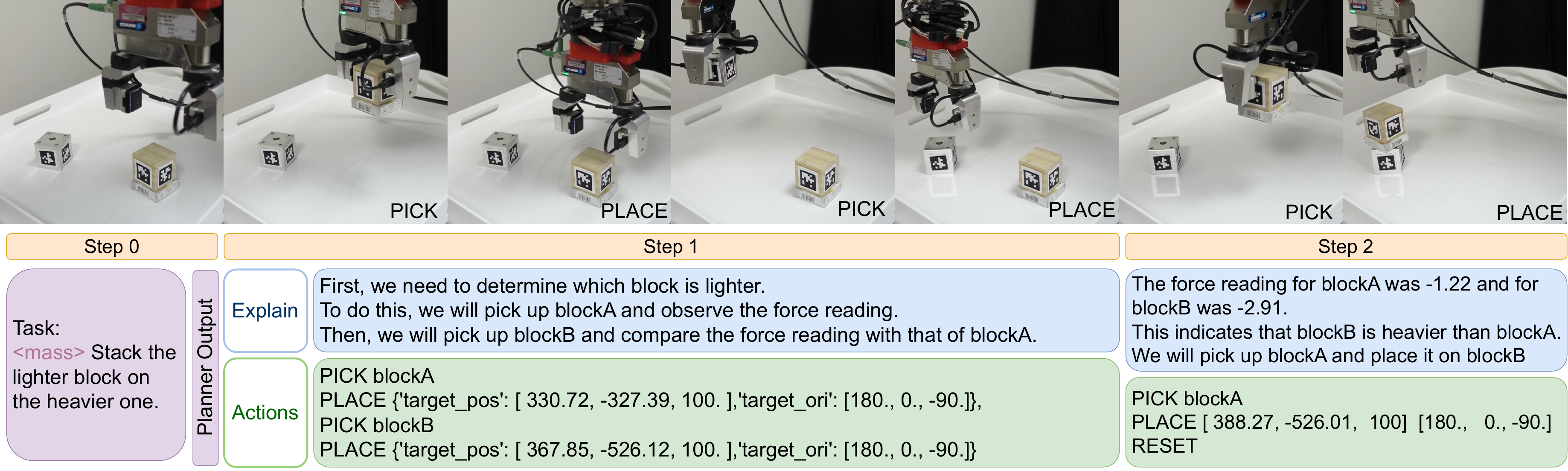}
    
    \caption{An example rollout of LLM-POP solving \textit{T4:Stack the lighter block on the heavier one}. In the first step, the LLM planner figured out the plan to determine which block is lighter by picking and placing up and placing down both blocks. In the second step, the LLM evaluator figured out blockB is heavier and plans to place blockA on blockB. In the next round(now shown in the figure), the evaluator recognized the completion of the task.}
    \label{fig:exp_demo}
\end{figure}
The experiments aim to validate the proposed interactive planner and to answer the following three questions:
\begin{enumerate}
    \item Is the proposed framework able to solve complex tasks with uncertainties?
    \item Does the framework apply to sim and real robots with various observation/action spaces?
    \item Can the fine-tuned LLM solve partial observable tasks? What are the gaps between GPT-4?
\end{enumerate}
\subsection{Experimental setup}
\noindent\textbf{Environment:} LLM-POP is evaluated on a set of manipulation tasks in a tabletop block rearrangement environment. A robot arm with a parallel gripper is equipped with a pre-trained skill set of \{Pick, Place, Reach, Reset\}. Each scenario is initialized with identical-size blocks with randomized positions and orientations on the table.

\noindent\textbf{Uncertainties:} We introduce two uncertainties in this environment 1) \textit{mass}: Density(mass) of the blocks is randomized. 2) \textit{fix}: blocks are randomized to be fixed/movable on the table. 
We design a task set containing tasks at different difficulty levels (horizon length to solve the task) under uncertainty assumptions to evaluate the planner's performance.

\noindent\textbf{Observations:}
The robot observations include the pose of the robot end effector, the pose of the blocks on the table, gripper opening positions, and force-torque (F/T) readings.  

\noindent\textbf{Pre-trained skills and parameters:} 
\emph{pick(object):} Pick up the specified \textit{object} on the table.
\emph{place(pose):} Move the end effector to desired \textit{pose} and open the gripper.
\emph{reach(pose):} Move the end effector to desired \textit{pose}.
\emph{reset():} Reset the arm and gripper to the initial pose.
where \textit{object} is a text name string, \textit{pose} is the position and orientation. 

\noindent\textbf{Evaluation metrics:} 
How to evaluate the task success rate is non-trivial, since the desired outcome of the tasks in Table~\ref{tab:tasks_set} (e.g., the lighter block is on top of the heavier block) could be achieved also through an incomplete decision-making process (e.g., stack a block on top of the other one that by chance respects the right weight relationship.) For this reason, even the LLM evaluator proposed in Section~\ref{sec:prompt_gpt4} cannot confidently determine the success rate in the tasks and we eventually relied on a manual check of each experiment


For each task in the evaluation task set, we evaluate the success rate of finishing the task under 10 random initializations on the positions and the uncertainties of blocks (5 for real robot settings since it's harder to randomize the uncertainties).
Table \ref{tab:tasks_set} includes the evaluation tasks we use. Default LLM-POP uses GPT-4 for the planner and evaluator.
\begin{table}[h]
\centering
\caption{Uncertainty and Task Description Table}
\begin{tabular}{l|l}
\Xhline{1pt}
Type & Task Descriptions \\
\hline
/ &        1. Stack one block onto another. \\
/ &        2. Move the blocks to the corners of the table. \\
mass &     3. Pick up the heavier block. \\
mass &     4. Stack the lighter block on the heavier. \\
fix &      5. Pick up the movable block and put it at the table corner. \\
fix &      6. Find the movable block and put it on the fixed block. \\
\Xhline{1pt}
\end{tabular}
\label{tab:tasks_set}
\vspace{-0.25in}
\end{table}

\subsection{Simulation: Block manipulation with Pre-trained LLM}
We first evaluate our approach for solving the tasks in Table \ref{tab:tasks_set} in a simulated robotic system in IsaacGym\cite{makoviychuk2021isaac}. We use the \textit{FrankaCubeStack} task as a template environment, but change the blocks to the same size with random densities for \textit{mass} uncertainty and randomly fix the block on the table for \textit{fix} uncertainty. The action parameters for \textit{place} include 3D target position and quaternion of the end effector. This is different from the block orientation quaternions in the observation, and explicitly including this (compared to using observation-action examples) in the prompt is essential for the LLM planner to get the correct action parameters and understand the observations. 

We use two ablations: 
1. GPT-3.5 as planner. 2. Remove the evaluator which explicitly asks for state abstraction and belief updates\footnote{To avoid the uncertainty from GPT versions, we use \textit{gpt-4-0314} for all GPT-4 and \textit{gpt-3.5-turbo-16k-0613} for GPT-3.5 usage.}. Evaluation results are shown in Table \ref{tab:simulation}.
\begin{table}[h]
\centering
    \begin{threeparttable}[b]
    \caption{Planner Success Rate on Evaluation Task Set}
    \begin{tabular}{lllllll}
\Xhline{1pt}
    Model     & Task 1     & Task 2 & Task 3 & Task 4 & Task 5 & Task 6 \\ 
    \hline
    GPT-3.5(w/o E)         &     5/10  &  7/10 &  0/10 & 0/10  & 0/10  & 0/10  \\
    GPT-3.5      &     6/10  &  6/10 &  0/10 & 0/10  & 0/10  & 0/10  \\
    LLM-POP(w/o E)           & 10/10 & 10/10  &  9/10 & 4/10  & 6/10  & 4/10  \\
    \textbf{LLM-POP}          &     \textbf{10/10}  &  \textbf{10/10} & \textbf{10/10}  & \textbf{8/10}  & \textbf{7/10}  & \textbf{8/10 } \\
\Xhline{1pt}
    LLM-POP*           & 5/5 & 5/5  & 5/5  & 4/5  & 5/5  & 4/5  \\
\Xhline{1pt}
    FT-Vanilla         & 4/10  & 2/10  & 0/10  & 0/10  & 0/10  &  0/10 \\
    FT-CoT             & 10/10 & 10/10 & 4/10  & 3/10  & 8/10 &   6/10 \\
\Xhline{1pt}
\end{tabular}
    \begin{tablenotes}  
       \item (w/o E): No evaluator. * Real-world experiment. FT: Fine-tuned Llama2 
     \end{tablenotes}
    \vspace{-0.1in}
    \label{tab:simulation}
    \end{threeparttable}
\end{table}

Compared to the GPT-4-based planner, GPT-3.5 is also able to reason correct action sequences for stacking tasks, but can not always reason about the geometric (position and orientation) parameters. It can understand what's the missing information for tasks with uncertainty, but fails to generate multi-step plans to collect and update the information. 

For the GPT-4 based planner, we observe an improvement in performance especially on longer-horizon tasks with uncertainty when the evaluator is added to the pipeline. An example is shown in Figure \ref{fig:exp_demo}.
The LLM is asked to explicitly what is missing and how such information can be analyzed from historical observations. 
This enforces the LLM to perform ``state abstraction'' and ``belief update'' before planning. 
In experiments, the GPT-4 planner without the evaluator sometimes makes the wrong plan by repeating the same collecting actions even if it already collected sufficient information. As the history of observations grows longer, the chance that LLM makes wrong reasoning also increases. 
This is partially due to the long-text handling challenge for the current GPT-4 version, and decomposing the reasoning tasks into an evaluator helps improve the stability.

\subsection{Hardware: Real robot Block Manipulation with GPT}
We implement the same version of our method on a MELFA Assista robot arm with a WSG 32 two-finger gripper. We put AprilTags~\cite{olson2011tags} on the sides of the blocks to get the pose estimate of blocks. We use blocks with different materials and added weight for uncertainty in mass. We use the force-torque (F/T) sensor mounted at the robot's wrist to get force readings. The action parameters for position controls on the real robot are in Euler angles and the angle for the gripper and blocks are in different frames. For safety, we set the task to fail if action parameters get out of safety bounds or a collision happens. Results are in Table \ref{tab:simulation}.

The LLM-POP framework (GPT-4) achieves better performance in the real robot compared to the simulation domain. This is mostly because of the very accurate position controller implemented on the real robot leading to fewer execution errors. The stiffness controller and F/T sensor used on the real robot allows us to recover accurate force readings with no noise compared to the ``sensor'' in the simulator which is affected by robot movement and gravity. These experiments show that the proposed framework can solve tasks with varying levels of difficulties and uncertainties reliably in simulated as well as real systems.


\subsection{Simulation: Block manipulation with fine-tuned model}

Using the self-instruction method introduced in \ref{sec:fine-tune}, we generate an Alpaca~\cite{alpaca}-like dataset and use it to finetune a Llama2-7B~\cite{touvron2023llama2} model for reasoning. We test it on the same IsaacGym environment. Results are also shown in Table \ref{tab:simulation}. 
\textit{FT-Vanilla} uses directly generated instruction pairs, while \textit{FT-CoT} uses CoT decomposition instruction pairs.
The biggest challenge during data generation for both is the correspondence between imagined observation generated by GPT-4 and the ground truth. For example, after picking up the block, the gripper position in the generated data is sometimes not close to the block position (and thus incorrect). This increases the difficulty for fine-tuned models to do correct reasoning based on observations. 
In the experiments, the fine-tuned model is able to reason about the missing information based on the task description and generate plans to collect the information. The results show that the fine-tuned model benefits from the CoT decomposition of instructions, and failures mostly come from wrong reasoning (wrong ``heavy'' block based on history). The current gap between the fine-tuned model and GPT-4 lies in the ability to analyze the historical information for updating the information and the ability to avoid and adjust wrong action parameters since it's not included in the current CoT design. A potential improvement is to add an auxiliary task of ``observation understanding'' in training and use diverse environment settings to improve the reasoning capability. We leave this to our future research.

\subsection{Common Failures in Sim and Real Experiments}
\subsubsection{Execution failures} This appears more in the simulation environment when the place action sets a target pose with less tolerance between objects and the robot moves at high speed (The control gains in the simulator are not fine-tuned for various block weights). LLM planners can also generate actions that cause collisions since there's no online collision avoidance in skills.

To explicitly test if the evaluator can help in correcting execution errors, we did an ablation on stacking(\textit{T1}) by adding offsets (1cm) on the grasping position of the block. The initial target placing position will fail because of this offset. With the evaluator included, which asks the LLM to analyze failure action based on history and propose correcting suggestions, the planner outputs a better target position (higher) in the next round. This shows that having an evaluator can actually help to correct execution errors. However, the execution success rate in this test also depends on the lower-level grasping skill; better grasp proposal~\cite{zhu2021grasp} and planning~\cite{wang2023simple} modules can improve the performance. Detailed analysis is deferred to a longer draft of the paper. 
\subsubsection{Belief update failures} Incorrect plan for collecting information (e.g., trying to reach above the block to measure its weight); wrong analysis results from the observations (e.g., not comparing weight using the force sensor z-axis value but using noise in other axes). In the LLM-POP with GPT-4 case, most failures come from the wrong analysis.



\section{Discussion}
In this work, we proposed an interactive planning framework LLM-POP using LLM to solve tasks under partial observation. The framework is verified in simulated as well as real robot systems on various partially observable tasks. 
Task distribution that the current framework can solve strongly depends on the diversity and robustness of the pre-trained skills. Current skills are open-loop actions based on initial observation. If the pre-trained skills are closed-loop policies with collision avoidance and online adjustment, the framework would be able to solve more challenging tasks. 

Overall, the gap between fine-tuned model and GPT-4 is clear, especially in reasoning for complex tasks. Our goal is not to replace the GPT-4 but to propose a method for generating self-instruct data for robotic tasks with limited demonstration data. We verify its usage as an interactive planner in manipulation and leave the task of learning more generalizable~\cite{sun2022mtrl} and transferrable sub-task policies~\cite{zhang2023efficient,sun2023mtrltransfer}, involving more modalities in the observation like image representations in robotics\cite{zhu2023minigpt, liu2023llava,huo2023humanoriented} to our future research. Similar settings can generalize to navigation settings and solving complex environments like HomerRobot~\cite{homerobotovmm} and social interactive scenarios~\cite{sun2023distributed}.

\bibliographystyle{abbrv}
\bibliography{neurips}

\appendix
\section{Appendix}
\subsection{Prompt example for GPT-4 evaluator}
\label{appendix:prompt_evaluator}
\begin{tcolorbox}[
colback=gray!20,
colframe=gray!60,
rounded corners,
title=Round 0 evaluator prompt for task: <mass> Stack the lighter block on the heavier block.
]
{[\textbf{Environment Context}]} \\A Franka Panda robot is placed in front of a table with cubes on top. The robot is equipped with a parallel gripper that can be used to pick and place the cubes. There is a force sensor on the end effector. The force sensor reading unit is Newton, gravity direction is -Z axis. Observations include the pose of the robot's end effector, the pose of the cubes, and the force readings. The positions of cubes are the center of mass position. The cube size is 0.05x0.05x0.05(m). The end effector with quaternion {[1 0 0 0]} is the pose of a finger pointing down to the ground. Note that the cube axis is different from the end effector axis. Quaternion {[0 0 0 1]} is the normal pose of the cube without rotation. The gripper is fully closed with joint angles {[0.04, 0.04]} and fully open with joint angles {[0.00, 0.00]}.\\If there is a <mass> tag in the task description, the cubes are randomized to be different weights. If there's a <movable> tag in the task description, the cubes are randomized to be fixed or movable.In the default setting, blocks are movable and have the same weight.\\
{[\textbf{Task Description}]} \\
 <mass> Stack the lighter block on the heavier block.\\
{[\textbf{History}]} None.\\\\
Based on the [Task description], [History], please first evaluate if the robot has successfully completed the task.\\
If the task is not finished, please explain:\\
1. Is there any missing information to solve the task? If [History] is included, also explain:\\
2. From the [History], what is the new task-related information we get? Show which action and observation you analyze the info from.\\
3. Are there any failed actions (observation after the action is not as expected)? Recognize the action failure and suggestions to change the action. Action failures usually happen when the plan does not leave enough clearance between the objects.\\
The response includes two lines: the first line is the boolean evaluation, the second line is the reason for the evaluation and the analysis based on the [Current Observation] and [History].\\
Example:\\
SUCCESS\\The robot has successfully completed the task.\\
FAIL\\Task is not finished. The robot needs to pick up cubeA and place it on top of cubeB.\\
FAIL\\Based on history the action PLACE failed. A potential solution is to change the target position to a higher position.
\end{tcolorbox}

\begin{tcolorbox}[
colback=gray!20,
colframe=gray!60,
rounded corners,
title=Round 0 evaluator response.
]
FAIL\\
The task is not finished. Missing information to solve the task: the weight of the cubes.
\end{tcolorbox}

\begin{tcolorbox}[
colback=gray!20,
colframe=gray!60,
rounded corners,
title=Round 1 evaluator response.
]
FAIL\\
Task is not finished. Based on the force sensor readings in Action 0 and Action 2, cubeA is lighter than cubeB. The robot needs to pick up cubeA and place it on top of cubeB.
\end{tcolorbox}
\subsection{Prompt example for GPT-4 planner}
\label{appendix:prompt_planner}
\begin{tcolorbox}
[breakable,
  colback=gray!20,
  colframe=gray!60,
  rounded corners,
  listing only,
  listing options={basicstyle=\ttfamily, breaklines=true},
  title=Round 1 planner prompt for task: <mass> Stack the lighter block on the heavier block.
]
 {[\textbf{Environment Context}]} \\A Franka Panda robot is placed in front of a table with cubes on top. The robot is equipped with a parallel gripper that can be used to pick and place the cubes. There is a force sensor on the end effector. The force sensor reading unit is Newton, gravity direction is -Z axis. Observations include the pose of the robot's end effector, the pose of the cubes, and the force readings. The positions of cubes are the center of mass position. The cube size is 0.05x0.05x0.05(m). The end effector with quaternion {[1 0 0 0]} is the pose of a finger pointing down to the ground. Note that the cube axis is different from the end effector axis. Quaternion {[0 0 0 1]} is the normal pose of the cube without rotation. The gripper is fully closed with joint angles {[0.04, 0.04]} and fully open with joint angles {[0.00, 0.00]}.\\If there is a <mass> tag in the task description, the cubes are randomized to be different weights. If there's a <movable> tag in the task description, the cubes are randomized to be fixed or movable. In the default setting, blocks are movable and have the same weight.\\
{[\textbf{Task Description}]} \\
 <mass> Stack the lighter block on the heavier block.\\
 {[\textbf{Action Options}]} \\
 1) REACH <location> // end effector move to a desired pose\\
 2) PICK <object> // pick up(reach and grasp) an object\\
 3) PLACE <location> // place the grasped object to a location\\
 4) RESET // reset the end effector to the initial position\\
 5) WAIT // do nothing when you need to stay still for a while\\
 <location> is a 7D vector, first 3 dimension position, following 4d quat orientation. For REACH, <location> is the desired end effector pose. For PLACE, <location> is the desired pose of the end effector (not the object) to release the object. For PICK <object> should be cubeA or cubeB. PICK includes a REACH action before grasping the object. No need to include REACH before PICK. PLACE should follow a PICK action. PLACE is needed after PICK if there's another PICK in the following actions. This is important for safety. Always note that the end effector pose is different from the object pose, and PLACE is the pose of the end effector with {[1 0 0 0]} quaternion pointing down.\\
 {[\textbf{Output Instruction}]}\\
 The output has two components: explanation and action sequence.\\Explanation start with EXPLAIN, then give words to explain your plan. For action sequence, you must first output EXECUTE, then give a sequence of actions and put each on a new line. Don't include extra explanations. Put exactly the single action on each line. Only include the action you are confident the robot should execute. Don't include the action if there are multiple choices and you are not certain. Plan it in the future with feedback.\\
 {[\textbf{Output Examples}(Not exactly for this task)]}\\EXPLAIN\\To move cubeA to the left of cubeB, pick up cubeA, and place it to the left of cubeB.\\EXECUTE\\ACTION PICK cubeA\\ACTION PLACE {[0. 0. 1.10 1 0 0 0]}\\ACTION RESET\\
 {[\textbf{History}]}\\
 == Round 0 ==\\{[Response History]}\\
 Plan explanation: First, I will pick up cubeA and lift it slightly to check its weight. Then, I will pick up cubeB and lift it slightly to check its weight. After comparing the weights, I will stack the lighter block on top of the heavier block.\\
 Action 0: PICK cubeA\\
 Observation after step 0: CubeA: {[0.11 0.13 1.20 0.01 0.00 0.39 0.92]} CubeB: {[0.18 -0.03 1.05 -0.00 0.00 -0.12 0.99]} EEF: {[0.10 0.13 1.20 0.93 0.37 -0.01 -0.00]} Force: {[1.17 -3.34 -3.95]} Gripper: {[0.02 0.02]}\\\\
 Action 1: PLACE {[0.10 0.13 1.10 1 0 0 0]}\\Observation after step 1: CubeA: {[0.12 0.13 1.05 -0.00 0.00 0.16 0.99]} CubeB: {[0.18 -0.03 1.05 -0.00 0.00 -0.12 0.99]} EEF: {[0.10 0.13 1.30 1.00 0.00 0.00 -0.00]} Force: {[-0.01 -0.00 0.01]} Gripper: {[0.04 0.04]}\\\\
 Action 2: RESET\\Observation after step 2: CubeA: {[0.12 0.13 1.05 -0.00 0.00 0.16 0.99]} CubeB: {[0.18 -0.03 1.05 -0.00 0.00 -0.12 0.99]} EEF: {[0.00 0.00 1.20 1.00 -0.00 -0.00 0.00]} Force: {[0.13 -0.57 -0.55]} Gripper: {[0.04 0.04]}\\\\
 Action 3: PICK cubeB\\Observation after step 3: CubeA: {[0.12 0.13 1.05 -0.00 0.00 0.16 0.99]} CubeB: {[0.18 -0.04 1.20 -0.00 0.00 -0.11 0.99]} EEF: {[0.18 -0.04 1.19 0.79 0.61 -0.01 -0.01]} Force: {[1.85 -5.00 -5.00]} Gripper: {[0.02 0.02]}\\\\
 Action 4: PLACE {[0.18 -0.03 1.10 1 0 0 0]}\\Observation after step 4: CubeA: {[0.12 0.13 1.05 -0.00 0.00 0.16 0.99]} CubeB: {[0.17 -0.03 1.05 -0.00 0.00 -0.67 0.74]} EEF: {[0.18 -0.03 1.30 1.00 0.00 0.00 -0.00]} Force: {[-0.01 0.00 0.00]} Gripper: {[0.04 0.04]}\\\\
 Action 5: RESET\\Observation after step 5: CubeA: {[0.12 0.13 1.05 -0.00 0.00 0.16 0.99]} CubeB: {[0.17 -0.03 1.05 -0.00 0.00 -0.67 0.74]} EEF: {[0.00 -0.00 1.20 1.00 0.00 -0.00 0.00]} Force: {[0.24 0.27 -0.89]} Gripper: {[0.04 0.04]}\\\\
 == Current Round ==\\{[\textbf{Current Observation} description]}\\Table surface center is at (0.00 0.00),Table surface size is (0.8 0.8)\\Cube A is at (0.12 0.13 1.05), with orientation (-0.00 0.00 0.16 0.99),Cube B is at (0.17 -0.03 1.05), with orientation (-0.00 0.00 -0.67 0.74), Robot State: End effector is at (0.00 -0.00 1.20), with orientation (1.00 0.00 -0.00 0.00), End effector hand force reading is (0.24 0.27 -0.89), Gripper position is 0.04, 0.04,\\
 {[\textbf{History and task evaluation}]}\\Task is not finished. Based on the force readings in Action 0 and Action 3, cubeA is lighter than cubeB. The robot needs to pick up cubeA and place it on top of cubeB.\\\\
 Based on {[History and task evaluation]}, and current observation, plan for the robot to collect information and solve the task in {[Task Description]}.\\Strictly follow the {[Action Options]} and {[Output Instruction]}.\\
\end{tcolorbox}

\begin{tcolorbox}[
colback=gray!20,
colframe=gray!60,
rounded corners,
title=Round 1 planner response.]
\textbf{EXPLAIN}\\
To stack the lighter block on the heavier block, I will pick up cubeA and place it on top of cubeB.\\
\textbf{EXECUTE}\\
ACTION PICK cubeA\\
ACTION PLACE [0.17 -0.03 1.10 1 0 0 0]
ACTION RESET\end{tcolorbox}
\newpage
\subsection{Prompt example for Task Generation}
\label{appendix:prompt_task}
\begin{tcolorbox}[
colback=gray!20,
colframe=gray!60,
rounded corners,
title=Prompt for task generation.]
You are serving as a task-generation helper for a given robot environment.

{[\textbf{General Environment Description}]} 
The environment is called "Block World". There is a 7DOF Franka Panda robot with a parallel gripper, it has a force sensor on the end effector. The robot is mounted on a table. There are multiple blocks on the table, here, we use 2 blocks as an example. The blocks are called {cubeA} and {cubeB}. The blocks are initialized at a random position on the table. The observation space (when there are two blocks) is a 26 dimension vector, consisting of:\\
cubeA position(3), cubeA orientation(4), cubeB position(3), cubeB orientation(4), end effector position(3), end effector orientation(4), end effector force(3), left finger position(1), right finger position(1)\\
If there are more than two blocks, the observation space is a 26+7*(n-2) dimension vector, where n is the number of blocks.\\
{[\textbf{Extra Environment Assumption Tags}]}\\
<weight> The blocks have randomized weights.\\
<move> The blocks are randomly determined to be movable or not; at least one block is movable.\\
<three> There are three blocks in the environment.\\
{[\textbf{Your Task}]}
Come up with 50 different tasks for the robot to perform. Each is designed under the assumption tags.\\
{[\textbf{Output format}]}
The response should follow the template below:
\#\#\# Task {i}: {task tag}{task description}\\
where {i} is the task number and {task description} is the task description.\\
The rules for task description:\\
1. Only include the objects in the environment in the task description.\\
2. The task description doesn't need to include all the objects in the environment.\\
3. The robot's basic skills are reach, grasp, and place. The task should not be out of its capability.\\
4. The task description can be implicit in the objects. For example, {Pick up the heavier block} is a valid task description.\\
5. The task description can be implicit in the goal. For example, {Maximize the height of the two blocks} is a valid task description.\\
6. Use your imagination to come up with different tasks. The tasks should be diverse and not too similar to each other.\\
7. You can include tasks with different levels of difficulty. Eazy tasks have short action sequences. Harder tasks have longer horizons which requires reasoning in planning.\\
8. Some tasks are not solvable with the initial observation. There are uncertainties in the task that require the robot to explore the environment to gather information. For tasks you think satisfy this requirement, please add a * at the end of the task description.\\
9. At least 30\% of the tasks should be non-solvable with the initial observation. \\
10. Tags can be combined together.\\
{[\textbf{Example}]}
Examples of {task tag}{task description}: \\
<move> find the movable cube and place it on top of the other block.*\\
<weight> move the heavier block to the corner of the table.* \\
<three><weight> sort all the blocks by their weight.* \\
<three> stack the three blocks. \\
\end{tcolorbox}
\newpage
\subsection{Prompt example for Instruction Generation}
\label{appendix:prompt_instruct}
\begin{tcolorbox}[
breakable,
colback=gray!20,
colframe=gray!60,
rounded corners,
title=prompt for CoT instruction generation]
You will be given a task in a robotic environment. You are asked to simulate the task instructions and corresponding responses happening during task solving. Some of them are long-horizon tasks request multiple reasoning steps, so we are generating multi-turn instructions in a chain of thought way. 
These task instructions will be given to a GPT model and we will evaluate the GPT model performance on the generated responses.\\
{[\textbf{General Environment Description}]}\\
<ENVIRONMENT DESCRIPTION PLACEHOLDER>\\
{[\textbf{Extra Environment Assumption Tags}]}\\
<TAGS PLACEHOLDER>\\
Tags at the beginning of {TASK} represent the environment assumptions for the task. In the default setting, blocks are movable and have the same weight.\\
{[\textbf{Instruction data Format}]}\\
The robot will be given a task: {TASK}. The instructions and responses happen when the robot is trying to solve this specific {TASK} and asks a chatbot guide. Each instruction data pair consists of three parts: {instruction}, {input}, {output}\\
The {instruction} consists of the question asked by the robot to help make decisions.\\
The {input} consists of the current observation and historical info. \\
The {output} consists of two parts <verbal> and <action>.\\
The <verbal> part describe the reasoning process and explanation for the current planned action if there is any.\\
The <action> part include a downstream action provided in the function lists executable by the robot. \\

The {instruction} of each task consists of the following standard questions in order to provide chain of thought instructions pairs.\\
1. Is the current information enough to solve the task? If not, what information is missing?\\
2. What are the actions the robot should take to gather information?\\
3. What are the actions the robot should take to solve the task?\\
For the 1st question, the <action> output part should be <nooutput>, only <verbal> output is important. The robot should ask this every time it collects new information. For the other questions, both <verbal> and <action> output parts are important. The 2nd and 3rd question usually happens when the answer to previous round question 1 is no(for question 2) or yes(for question 3).\\
When generating instruction data, you need to imagine the observation and previously collected information for the robot when asking the question and generate the corresponding {input}. The generated {output} should correspond to the {input} you created.\\
{[\textbf{Format of generated instructions}]}\\
1. The i-th response need to satisfy the following format. \\
// start of instruction pair i, not including this line.\\
\#\#\#\\
i.\\
<Task> {task}\\
<Instruction> {instruction}\\
<Input> {input}\\
<Output>\\
{[verbal]} {verbal output}\\
{[action]} {list of function output}\\
// end of instruction pair i, not including this line.\\
2. The index of instructions starts from 1.\\
3. The format of {instruction}: It's usually one of the questions listed above.\\
4. The format of {input} will be a vector of robot observation, followed by a list of historical information. Use actual numbers in the vectors. The format is: \\
Current: {[observation]}\\
Past:\\
Round 1: {[hist text 1]} {[hist action list 1]} {[hist obs list 1]}\\
Round 2: {[hist text 2]} {[hist action list 2]} {[hist obs list 2]}\\
...\\
{[hist text]} {[hist action list]} are the previous rounds explanation and action sequence, [hist obs list] is the observation after the action executions in previous rounds. The number of hist obs should correspond to the number of actions in hist action list. \\
5. The format of {verbal output} will be a sentence explain the current reasoning process and the current planned action.\\
6. The format of {action output} will be list of {function name} {function parameter} wrapped by []. Each element should be in a python executable form, don't use placeholders as parameters, output the numbers if the parameters are vectors.\\
7. Each instruction pair should be separated by a line of "\#\#\#" at the beginning.\\
8. Generate the multiple rounds of instructions from initial until the task is solved, the observations in history should be consistent with the observations in the current round.\\
9. When the task is finished, start from the beginning and solve the tasks from different initial states. Continue the index from the previous set of instructions. Repeat for at least 2 times.\\
10. There can be different responses to the same question depending on different {input}, when there's multiple potential input output pairs, try to cover different cases during the repeating.\\
{[\textbf{Tasks to solve}]}\\
Below is the {TASK} used in the generated instructions:\\
<TASK PLACEHOLDER>

{[\textbf{Function Lists in skill library}]}\\
<FUNCTION LIST PLACEHOLDER>

{[\textbf{Example observations}]}\\
Below are some example observations when executing some skills. Use them as a reference for imagined observations when generating instructions.\\
<TRAJECTORY PLACEHOLDER>

{[\textbf{Example instruction pairs}]}\\
Below are some example instruction pairs. Use them as a reference when generating instructions.\\
<INSTRUCTION EXAMPLE PLACEHOLDER>\\
Ignore the index in the example and start from 1.

\end{tcolorbox}
\newpage
\subsection{Examples of task generated}
\label{appendix:example_task}
\begin{tcolorbox}[
colback=blue!10,
colframe=blue!30,
rounded corners,
title=Example tasks generated.]
<weight> Pick up the heavier block*    \\
<move> Find the movable block and place it on top of the other block*\\
<weight> Move the heavier block to the corner of the table*\\
<three><weight> Sort all the blocks by their weight*\\
<three> Stack the three blocks\\
<weight> Pick up the lighter block*\\
<three> Move the heaviest block to the left corner of the table, and the lightest block to the right corner*\\
<move> Pick up the only movable block and hold it for 5 seconds*\\
<weight><move> Pick up the movable and lighter block*\\
<three> Stack two of the blocks together\\
<three> Place the blocks in a straight line with equal distance between them\\
<three><weight> Stack the blocks in order of their weight, with the heaviest at the bottom*\\
<move> Find a block that is both movable and lighter, if possible, and place it on top of the heavier block*\\
<weight> Hold the heavier block in the gripper for 10 seconds\\
<three><weight> Pick up the block with the medium weight
\end{tcolorbox}

\newpage
\subsection{Examples of instruction generated}
\label{appendix:example_instruct}
\begin{tcolorbox}[
breakable,
colback=blue!10,
colframe=blue!30,
rounded corners,
title=Example instruction pairs generated for selected tasks.]
 {
        1.\\
        \textcolor{blue}{<instruction>}: "Environment: Franka Panda robot with blocks on the table.\textbackslash nTask: <move> Find the movable block and place it on top of the other block*\textbackslash nIs the current information enough to solve the task? If not, what information is missing?\textbackslash n"\\
        \textcolor{blue}{<input>}: "Current:\textbackslash n{[0.27, 0.12, 1.025, 0, 0, 0, 1, -0.12, 0.15, 1.025, 0, 0, 0, 1, 0.0, 0.0, 1.50, 0, 0, 0, 1, 0, -1, 0, 0.0, 0.0]}\textbackslash nPast:\textbackslash n{[]} {[]} {[]}\textbackslash n"\\
        \textcolor{blue}{<output>}: "EXPLAIN: No. The robot doesn't know whether cubeA or cubeB is movable. We need to try moving both blocks.\textbackslash nACTION: <nooutput>."
    },\\
    2.\\
    {
        \textcolor{blue}{<instruction>}: "Environment: Franka Panda robot with blocks on the table.\textbackslash nTask: <move> Find the movable block and place it on top of the other block*\textbackslash nWhat are the actions the robot should take to gather information?\textbackslash n"\\
        \textcolor{blue}{<input>}: "Current:\textbackslash n{[0.27, 0.12, 1.025, 0, 0, 0, 1, -0.12, 0.15, 1.025, 0, 0, 0, 1, 0.0, 0.0, 1.50, 0, 0, 0, 1, 0, -1, 0, 0.0, 0.0]}\textbackslash nPast:\textbackslash nRound 1: {[No. The robot doesn't know whether cubeA or cubeB is movable. We need to try moving both blocks.]} {[]} {[]}\textbackslash n"\\
        \textcolor{blue}{<output>}: "EXPLAIN: The robot will try to grasp cubeA first and then cubeB in order to find the movable block.\textbackslash nACTION: {[grasp('cubeA'), grasp('cubeB')]}."
    },\\
    3.\\
    {
        \textcolor{blue}{<instruction>}: "Environment: Franka Panda robot with blocks on the table.\textbackslash nTask: <move> Find the movable block and place it on top of the other block*\textbackslash nIs the current information enough to solve the task? If not, what information is missing?\textbackslash n"\\
        \textcolor{blue}{<input>}: "Current:\textbackslash n{[-0.12, 0.15, 1.025, 0, 0, 0, 1, -0.12, 0.15, 1.2, 0, 0, 0, 1, -0.12, 0.15, 1.2, 1, 0, 0, 0, -1, 0, 0, 0.04, 0.04]}\textbackslash nPast:\textbackslash nRound 1: {[No. The robot doesn't know whether cubeA or cubeB is movable. We need to try moving both blocks.]} {[]} {[]}\textbackslash nRound 2: {[The robot will try to grasp cubeA first and then cubeB in order to find the movable block.]} {[grasp('cubeA'), grasp('cubeB')]} {[{[0.27, 0.12, 1.2, -0.02, 0.02, -0.14, 0.99, -0.12, 0.15, 1.025, 0, 0, 0, 1, 0.27, 0.12, 1.2, 0.8, 0.6, 0, 0, -1, -1.8, 0, 0.04, 0.04]}, {[-0.12, 0.15, 1.025, 0, 0, 0, 1, -0.12, 0.15, 1.2, 0, 0, 0, 1, -0.12, 0.15, 1.2, 1, 0, 0, 0, -1, 0, 0, 0.04, 0.04]}]}\textbackslash n"\\
        \textcolor{blue}{<output>}: "EXPLAIN: Yes. Block B is movable as the end effector moved the block when trying to grasp it.\textbackslash nACTION: <nooutput>."
    },\\
    4.\\
    {
        \textcolor{blue}{<instruction>}: "Environment: Franka Panda robot with blocks on the table.\textbackslash nTask: <move> Find the movable block and place it on top of the other block*\textbackslash nWhat are the actions the robot should take to solve the task?\textbackslash n"\\
        \textcolor{blue}{<input>}: "Current:\textbackslash n{[-0.12, 0.15, 1.025, 0, 0, 0, 1, -0.12, 0.15, 1.2, 0, 0, 0, 1, -0.12, 0.15, 1.2, 1, 0, 0, 0, -1, 0, 0, 0.04, 0.04]}\textbackslash nPast:\textbackslash nRound 1: {[No. The robot doesn't know whether cubeA or cubeB is movable. We need to try moving both blocks.]} {[]} {[]}\textbackslash nRound 2: {[The robot will try to grasp cubeA first and then cubeB in order to find the movable block.]} {[grasp('cubeA'), grasp('cubeB')]} {[{[0.27, 0.12, 1.2, -0.02, 0.02, -0.14, 0.99, -0.12, 0.15, 1.025, 0, 0, 0, 1, 0.27, 0.12, 1.2, 0.8, 0.6, 0, 0, -1, -1.8, 0, 0.04, 0.04]}, {[-0.12, 0.15, 1.025, 0, 0, 0, 1, -0.12, 0.15, 1.2, 0, 0, 0, 1, -0.12, 0.15, 1.2, 1, 0, 0, 0, -1, 0, 0, 0.04, 0.04]}]}\textbackslash nRound 3: {[Yes. Block B is movable as the end effector moved the block when trying to grasp it.]} {[]} {[]}\textbackslash n"\\
        \textcolor{blue}{<output>}: "EXPLAIN: The robot should grasp the movable cubeB and place it on the top of cubeA.\textbackslash nACTION: {[grasp('cubeB'), place('cubeB', {[0.27, 0.12, 1.075]}, {[0, 0, 0, 1]})]}.\textbackslash n\textbackslash n\textbackslash n{[Repeat from different initial state]}\textbackslash nRepeat the above instructions from a different initial observation vector."
    },\\
    5.\\
    {
        \textcolor{blue}{<instruction>}: "Environment: Franka Panda robot with blocks on the table.\textbackslash nTask: <move> Find the movable block and place it on top of the other block*\textbackslash nIs the current information enough to solve the task? If not, what information is missing?\textbackslash n"\\
        \textcolor{blue}{<input>}: "Current:\textbackslash n{[-0.25, 0.10, 1.025, 0, 0, 0, 1, 0.27, -0.17, 1.025, 0, 0, 0, 1, 0.0, 0.0, 1.50, 0, 0, 0, 1, 0, 1, 0, 0.0, 0.0]}\textbackslash nPast:\textbackslash n{[]} {[]} {[]}\textbackslash n"\\
        \textcolor{blue}{<output>}: "EXPLAIN: No. The robot doesn't know whether cubeA or cubeB is movable. We need to try moving both blocks.\textbackslash nACTION: <nooutput>."
    },\\
    6.\\
    {
        \textcolor{blue}{<instruction>}: "Environment: Franka Panda robot with blocks on the table.\textbackslash nTask: <move> Find the movable block and place it on top of the other block*\textbackslash nWhat are the actions the robot should take to gather information?\textbackslash n"\\
        \textcolor{blue}{<input>}: "Current:\textbackslash n{[-0.25, 0.10, 1.025, 0, 0, 0, 1, 0.27, -0.17, 1.025, 0, 0, 0, 1, 0.0, 0.0, 1.50, 0, 0, 0, 1, 0, 1, 0, 0.0, 0.0]}\textbackslash nPast:\textbackslash nRound 5: {[No. The robot doesn't know whether cubeA or cubeB is movable. We need to try moving both blocks.]} {[]} {[]}\textbackslash n"\\
        \textcolor{blue}{<output>}: "EXPLAIN: The robot will try to grasp cubeA first and then cubeB in order to find the movable block.\textbackslash nACTION: {[grasp('cubeA'), grasp('cubeB')]}."
    },\\
    7.\\
    {
        \textcolor{blue}{<instruction>}: "Environment: Franka Panda robot with blocks on the table.\textbackslash nTask: <move> Find the movable block and place it on top of the other block*\textbackslash nIs the current information enough to solve the task? If not, what information is missing?\textbackslash n"\\
        \textcolor{blue}{<input>}: "Current:\textbackslash n{[-0.25, 0.10, 1.2, -0.02, 0.02, -0.14, 0.99, 0.27, -0.17, 1.025, 0, 0, 0, 1, -0.25, 0.10, 1.2, 0.8, 0.6, 0, 0, 0.96, -0.18, -1.8, 0.02, 0.02]}\textbackslash nPast:\textbackslash nRound 5: {[No. The robot doesn't know whether cubeA or cubeB is movable. We need to try moving both blocks.]} {[]} {[]}\textbackslash nRound 6: {[The robot will try to grasp cubeA first and then cubeB in order to find the movable block.]} {[grasp('cubeA'), grasp('cubeB')]} {[{[-0.25, 0.10, 1.2, -0.02, 0.02, -0.14, 0.99, 0.27, -0.17, 1.025, 0, 0, 0, 1, -0.25, 0.10, 1.2, 0.8, 0.6, 0, 0, 0.96, -0.18, -1.8, 0.02, 0.02]}]}\textbackslash n"\\
        \textcolor{blue}{<output>}: "EXPLAIN: Yes. Block A is movable as the end effector moved the block when trying to grasp it.\textbackslash nACTION: <nooutput>."
    },\\
    8.\\
    {
        \textcolor{blue}{<instruction>}: "Environment: Franka Panda robot with blocks on the table.\textbackslash nTask: <move> Find the movable block and place it on top of the other block*\textbackslash nWhat are the actions the robot should take to solve the task?\textbackslash n"\\
        \textcolor{blue}{<input>}: "Current:\textbackslash n{[-0.25, 0.10, 1.2, -0.02, 0.02, -0.14, 0.99, 0.27, -0.17, 1.025, 0, 0, 0, 1, -0.25, 0.10, 1.2, 0.8, 0.6, 0, 0, 0.96, -0.18, -1.8, 0.02, 0.02]}\textbackslash nPast:\textbackslash nRound 5: {[No. The robot doesn't know whether cubeA or cubeB is movable. We need to try moving both blocks.]} {[]} {[]}\textbackslash nRound 6: {[The robot will try to grasp cubeA first and then cubeB in order to find the movable block.]} {[grasp('cubeA'), grasp('cubeB')]} {[{[-0.25, 0.10, 1.2, -0.02, 0.02, -0.14, 0.99, 0.27, -0.17, 1.025, 0, 0, 0, 1, -0.25, 0.10, 1.2, 0.8, 0.6, 0, 0, 0.96, -0.18, -1.8, 0.02, 0.02]}]}\textbackslash nRound 7: {[Yes. Block A is movable as the end effector moved the block when trying to grasp it.]} {[]} {[]}\textbackslash n"\\
        \textcolor{blue}{<output>}: "EXPLAIN: The robot should grasp the movable cubeA and place it on the top of cubeB.\textbackslash nACTION: {[grasp('cubeA'), place('cubeA', {[0.27, -0.17, 1.075]}, {[0, 0, 0, 1]})]}."
    }
\end{tcolorbox}

\end{document}